\pdfoutput=1

\documentclass[sigconf]{acmart}

\AtBeginDocument{%
  \providecommand\BibTeX{{%
    \normalfont B\kern-0.5em{\scshape i\kern-0.25em b}\kern-0.8em\TeX}}}

\copyrightyear{2022}
\acmYear{2022}
\setcopyright{rightsretained}
\acmConference[FAccT '22]{2022 ACM Conference on Fairness, Accountability, and Transparency}{June 21--24, 2022}{Seoul, Republic of Korea}
\acmBooktitle{2022 ACM Conference on Fairness, Accountability, and Transparency (FAccT '22), June 21--24, 2022, Seoul, Republic of Korea}\acmDOI{10.1145/3531146.3533154} \acmISBN{978-1-4503-9352-2/22/06}

\acmConference[FAccT '22]{FAccT '22: ACM Conference on Fairness, Accountability, and Transparency}{June 21--24, 2022}{Seoul, South Korea}
\acmBooktitle{FAccT '22: ACM Conference on Fairness, Accountability, and Transparency,
  June 21--24, 2022, Seoul, South Korea}
\acmPrice{15.00}
\acmISBN{978-1-4503-XXXX-X/18/06}

\acmSubmissionID{5766027}

\usepackage{microtype}
\usepackage{booktabs}
\usepackage{multirow}
\usepackage{graphicx}
\usepackage[inline]{enumitem}
\usepackage{fancybox}
\usepackage{multicol}
\usepackage[most]{tcolorbox}
\AtBeginDocument{
\tcbset{enlarge top initially by=0mm,enlarge bottom finally by=0mm, boxsep=0mm,left=1mm,right=1mm,breakable,enhanced,boxrule=0.3mm,fontlower=\small,fontupper=\small}}

\newcommand{\xhdr}[1]{\textbf{#1.}\ \ }

\begin{document}

\title{Multi-disciplinary fairness considerations in machine learning for clinical trials}

\author{Isabel Chien}
\email{ic390@cam.ac.uk}
\orcid{0000-0001-7207-8526}
\affiliation{
  \institution{University of Cambridge}
  \streetaddress{James Dyson Building}
  \city{Cambridge}
  \country{UK}
  \postcode{CB2 1QA}
}
\author{Nina Deliu}
\email{nina.deliu@mrc-bsu.cam.ac.uk}
\orcid{0000-0003-2501-8795}
\affiliation{
  \institution{MRC Biostatistics Unit, University of Cambridge}
  \city{Cambridge}
  \country{UK}
}
\affiliation{
  \institution{MEMOTEF, Sapienza University of Rome}
  \city{Rome}
  \country{IT}
}

\author{Richard E. Turner}
\email{ret26@cam.ac.uk}
\orcid{0000-0003-0066-0984}
\affiliation{
  \institution{University of Cambridge}
  \city{Cambridge}
  \country{UK}
}

\author{Adrian Weller}
\email{aw665@cam.ac.uk}
\orcid{0000-0003-1915-7158}
\affiliation{
  \institution{University of Cambridge}
  \city{Cambridge}
  \country{UK}
}
\affiliation{
  \institution{The Alan Turing Institute}
  \city{London}
  \country{UK}
}

\author{Sofia S. Villar}
\email{sofia.villar@mrc-bsu.cam.ac.uk}
\orcid{0000-0001-7755-2637}
\affiliation{
  \institution{MRC Biostatistics Unit, University of Cambridge}
  \city{Cambridge}
  \country{UK}
}

\author{Niki Kilbertus}
\email{niki.kilbertus@tum.de}
\orcid{1234-5678-9012}
\affiliation{
  \institution{Technical University of Munich}
  \city{Munich}
  \country{Germany}
}
\affiliation{
  \institution{Helmholtz Munich}
  \city{Munich}
  \country{Germany}
}
\renewcommand{\shortauthors}{Chien, et al.}

\begin{abstract}
While interest in the application of machine learning to improve healthcare has grown tremendously in recent years, a number of barriers prevent deployment in medical practice. A notable concern is the potential to exacerbate entrenched biases and existing health disparities in society. The area of fairness in machine learning seeks to address these issues of equity; however, appropriate approaches are context-dependent, necessitating domain-specific consideration. We focus on clinical trials, i.e., research studies conducted on humans to evaluate medical treatments. Clinical trials are a relatively under-explored application in machine learning for healthcare, in part due to complex ethical, legal, and regulatory requirements and high costs. Our aim is to provide a multi-disciplinary assessment of how fairness for machine learning fits into the context of clinical trials research and practice. We start by reviewing the current ethical considerations and guidelines for clinical trials and examine their relationship with common definitions of fairness in machine learning. We examine potential sources of unfairness in clinical trials, providing concrete examples, and discuss the role machine learning might play in either mitigating potential biases or exacerbating them when applied without care. Particular focus is given to adaptive clinical trials, which may employ machine learning. Finally, we highlight concepts that require further investigation and development, and emphasize new approaches to fairness that may be relevant to the design of clinical trials.
\end{abstract}



\maketitle

\section{Introduction}
Recent advancements in the field of machine learning (ML) have fueled considerable excitement regarding its potential to change the landscape of healthcare and biomedical research globally~\citep{obermeyer2016predicting,rajkomar2019machine}. While much of current research has focused on the application of ML to benefit clinical practice, ML may also offer improvements to clinical research~\citep{Weissler2021TheRO, Woo2019AnAB}. Clinical research refers to the study of the safety and efficacy of medical, surgical, or behavioral interventions that are intended for human use~\citep{Bothwell2017TheRE}. A common tool of clinical research is the clinical trial (CT): experiments conducted to evaluate the impact of treatments or interventions on human subjects, ranging from early-phase dose-finding trials to confirmatory studies. Randomized controlled trials (RCTs) are considered the ``gold standard'' for evaluating the effectiveness of experimental interventions~\citep{stolberg2004randomized, akobeng2005understanding}, due to their strong statistical guarantees~\citep{Bothwell2017TheRE,rosenberger2019randomization}. However, the traditional fixed design allows no flexibility for beneficial alterations. As a result, interest is growing in adaptive designs (ADs), wherein results from an ongoing clinical trial may be incorporated into trial procedure changes~\citep{pallmann2018adaptive, chow}.

Researchers and commercial entities have proposed ML-driven improvements to the design of clinical trials, but extensive research and adoption in practice have been limited~\citep{Weissler2021TheRO}. A likely explanation is that clinical trials are subject to a complex mixture of ethical, legal, and regulatory considerations, and, as yet, there exists limited guidance for the inclusion of ML~\citep{WawiraGichoya2021EquityIE,Weissler2021TheRO}. The design of clinical trials remains the subject of ongoing ethical discussions, with some criticizing the morality of RCTs~\citep{Royall1991EthicsAS,Nardini2014TheEO}, and others detailing possible issues surrounding adaptive designs~\citep{vanderGraaf2012AdaptiveTI,Bothwell2017TheRE,Park2018CriticalCI}. To complicate matters further, ethical principles currently considered when assessing clinical research studies may be insufficient to address additional concerns stemming from the inclusion of ML. 

Nevertheless, ML has been proposed to assist with clinical trials planning, participant management, data collection, and analysis~\citep{Weissler2021TheRO}. Deep learning models have been developed for patient-trial matching, using electronic health records (EHR) data and trial eligibility criteria to recommend suitable patients for certain trials~\citep{Hassanzadeh2020MatchingPT, Zhang2020DeepEnrollPM}. Such systems have performed well in practice; a study found that the IBM Watson for Clinical Trial matching system, used to match breast cancer patients with systemic therapy trials, increased average monthly enrollment by $80\%$ over the 18 months following implementation~\citep{Haddad2018ImpactOA}. From a more theoretical perspective, reinforcement learning (RL) and multi-armed bandit (MAB) methods have been considered for modelling safe, effective doses in adaptive dose-finding trials ~\citep{shen2020learning,lee2020contextual,aziz2021multi}.

While ML has potential to improve the welfare and fairness across participants in clinical trials, there is a risk that ML exacerbates existing inequalities or introduces new disparities and biases~\citep{ferryman,paulus,wiens}. ML model outputs are dependent on their training datasets, which may reflect historical biases~\citep{rajkomar}, thus perpetuating existent disparities in society. Women and racial/ethnic minorities are systematically underrepresented in clinical research, contributing to disproportionately higher incidence of and mortality from cancer~\citep{desantis} and cardiovascular diseases~\citep{cho2021increasing}, and public health misconceptions~\citep{bradley2021unrepresentative}. Healthcare algorithms used in practice as well as those considered state-of-the-art for certain clinical prediction tasks have been discovered to exhibit racial~\citep{obermeyer}, gender, and socioeconomic biases~\citep{chen2019}.

Fairness in machine learning addresses disparities between sub-populations in data-driven decision-making systems to mitigate inequalities. However, the intersection of machine learning, fairness, and healthcare is still relatively under-explored, with many unresolved ethical questions~\citep{char2018}. Existing fairness criteria have been found to be mutually incompatible~\citep{paulus} and conceptions of fairness addressing exploration-exploitation trade-offs require more attention \citep{chouldechova}. Approaches to and definitions of fairness are context-dependent, requiring domain-specific consideration. An interdisciplinary approach including ethicists, medical practitioners, social and other quantitative scientists is required to assess appropriate formal fairness measures, interventions, and necessary trade-offs~\citep{wiens, rajkomar}. Bias can arise at every step of the process, including study design, data collection, data analysis, model building, model evaluation, and even deployment~\citep{rajkomar, char2018, chen2021}.
Researchers increasingly emphasize the need for formal rules and regulations that explicitly address issues of fairness in ML for healthcare~\citep{WawiraGichoya2021EquityIE}. 

In this paper, we aim to contextualize fairness considerations within ML for clinical trials, to provide structure and clarity for those hoping to leverage ML to improve the practice of clinical trials. We consider ML from several angles: its potential to improve patient welfare and equity, the risks of propagating unfairness and adverse effects, and its potential to facilitate the use of adaptive designs, which could be considered fairer than RCTs. We highlight the following contributions:
\begin{enumerate}[leftmargin=*,topsep=0pt,itemsep=0pt]
    \item After a primer on the design and implementation of clinical trials, we discuss the surrounding ethical considerations and guidelines from an ML perspective and establish relationships with notions of fairness in ML.
    \item We examine potential sources of unfairness in clinical trials alongside concrete examples to illustrate where adaptive designs or ML can be used to facilitate improvements.  
    \item We curate a list of opportunities for innovation in ML to address sources of unfairness, providing an overview of current literature and highlighting critical directions for future work. 
\end{enumerate}

\section{Background: Clinical Trials, Ethics, and Fairness}
In this section, we provide background for the multi-disciplinary discussion of ethical usage of ML in clinical trials. We characterize the modern practice of clinical trials, from design to implementation. Following that, we review influential documents and regulations that govern clinical research alongside renewed contemporary ethical discussions. Finally, we briefly describe ongoing approaches to and open questions within fairness in ML.

\subsection{Structure of Clinical Trials}
\subsubsection{Clinical Trial Designs}
\label{designs}
CTs are research studies in which participants are allocated to different treatments in order to evaluate the effectiveness and safety of treatments of interest. They are commonly classified into four phases: dose-finding phase-I trials, testing for safety in a few human subjects; phase-II efficacy evaluation trials; confirmatory phase-III trials, testing for effectiveness; and phase-IV surveillance studies, for long-term effects monitoring~\citep{fda_clin_research}. 

Patient allocation may be randomized, where each participant has a known positive probability of being assigned to a treatment arm, or non-randomized~\citep{kalish1985treatment} with rule-based approaches~\citep{wheeler2019design}. Well-designed RCTs are recognized as the ``gold standard'' for conducting confirmatory evidence-based evaluation of experimental interventions~\citep{stolberg2004randomized, akobeng2005understanding}. In a RCT, participants are randomly allocated to either control arms (standard of care or placebo), or experimental arms~\citep{Kendall2003DesigningAR}. Randomization removes the influence of potential confounders on study outcomes~\citep{Saturni2014RandomizedCT, Stanley2007DesignOR}, mitigating bias and enabling valid cause-effect estimation~\citep{rosenberger2019randomization}. Traditional RCTs are static; key elements (e.g., sample size, randomization probabilities) remain fixed during the course of the trial, with the exception that a trial may be stopped due to significant safety concerns. This may be limiting, as interim analysis may provide enough evidence for stopping the trial earlier for success, lack of efficacy, or skewing allocation towards the most beneficial treatment. 

In adaptive designs, key trial characteristics (e.g., treatment doses or allocation probabilities) may be altered during the course of the study based on accumulating data (e.g., treatment responses), according to predefined rules~\citep{pallmann2018adaptive, chow}. Examples include: (i) \textit{response-adaptive randomization} (RAR), where allocation ratios may be shifted towards more promising or informative treatments~\citep{pallmann2018adaptive}; (ii) \textit{drop the loser}, where inferior treatment arms may be removed~\citep{burnett2020adding,pallmann2018adaptive}; or (iii) \textit{adaptive dose-finding}, where safe (maximum tolerated) dosages are determined according to rule- or model-based strategies~\citep{wheeler2019design,burnett2020adding}. Multiple adaptations may also be incorporated into a single trial. While adaptive designs can be more efficient, informative, and ethical than traditional RCTs, as they are able to adjust to new information and make better use of limited resources~\citep{pallmann2018adaptive,chow}, they may create new ethical issues, increase costs, and complicate statistical analysis. For a detailed overview on adaptive designs, we refer the interested reader to~\citep{bauer2016twenty,pallmann2018adaptive,bhatt2016adaptive}.

While public interest in adaptive designs seems to rise during healthcare crises, uptake in practice has been minimal, particularly in relation to the methodological/theoretical literature~\citep{robertson,Sverdlov2021OpportunityFE}. Common concerns about adaptive designs stem from lack of clarity on the methodology involved in running such trials and applicability of traditional statistical inference methods to interpret trial outcomes~\citep{pallmann2018adaptive}. In recent years, proponents of adaptive designs have sought to guide understanding of how they can be successfully implemented and what they can achieve~\citep{Angus2019AdaptivePT,pallmann2018adaptive,chow,robertson,burnett2020adding}, and to pursue developments in machine learning that may support adaptive methods~\citep{villar2015,Villar2015ResponseadaptiveRF,williams2021,williamson,aziz2021multi,Bastani2015OnlineDW,Lee2021SDFBayesCO}.

\subsubsection{Clinical Trials in Practice}
Proposed CTs are typically outlined in a \textit{clinical trial protocol}, which must then be submitted to regulatory bodies and ethics committees for review and approval before trial implementation~\citep{Chan2013SPIRIT2S}. It is a document written by study investigators that delineates the motivations of a trial and the planned methodologies by which a trial will be conducted~\citep{Chan2013SPIRIT2S,Rivera2020GuidelinesFC}. To help standardize trial protocols, an international committee published the SPIRIT (Standard Protocol Items: Recommendations for Interventional Trials) guidelines (2013)~\citep{Chan2013SPIRIT2S}, as well as a SPIRIT-AI Extension (2020), which addresses interventions that involve AI~\citep{Rivera2020GuidelinesFC}. The SPIRIT guidelines recommend that a trial protocol include comprehensive details such as (but not limited to) justification for undertaking the trial, eligibility criteria, interventions being studied, safety considerations, trial design, and the statistical analysis plan.


The regulatory bodies and ethics committees responsible for trial approval vary by country. In the United States (US), the Food and Drug Administration (FDA) is responsible for regulatory approval~\citep{fda_clin_research} and Institutional Review Boards independently determine ethical approval~\citep{fda_irb}. In the United Kingdom, trial protocols are submitted to multiple agencies for various ethics and regulatory approvals~\citep{ukri}. Trials are assessed based on the aim of the trial, whether likely benefits outweigh risks, the planned design and analysis of the trial, among other aspects~\citep{fda_clin_research,cancer_research_uk}. Treatments are typically approved by regulatory bodies following successful confirmatory phase-III trials~\citep{fda_clin_research}.


\subsection{Ethics of Clinical Trials}
\subsubsection{Current Ethical Guidelines}
\label{ethicalguidelines}
Ethical guidelines regarding the conduct of clinical research have been developed for guidance against harming or exploiting patient volunteers, while preserving the integrity of the scientific research. \citet{Emanuel2000WhatMC} have developed a widely-cited framework consisting of seven requirements for assessing the ethics of clinical research studies. We take guidance from these seven requirements~\citep{Emanuel2000WhatMC,nihethics}, with further discussion in Appendix~\ref{sec:app:ethicalprinciples}:
\begin{enumerate*}
    \item value for society,
    \item scientific validity,
    \item fair subject selection,
    \item favorable risk-benefit ratio,
    \item independent review,
    \item informed consent, and
    \item respect for enrolled subjects.
\end{enumerate*}
However, these guidelines do not sufficiently address issues arising from the growing uptake of novel trial designs or the use of ML in clinical trials settings, which will inevitably bring about risks and burdens that should be explicitly acknowledged in any trial planning process.

\subsubsection{Guidelines for AI/ML in Clinical Trials}
Currently, there exists limited guidance on the ethical inclusion of ML in clinical research. Fairness is a key consideration in ethical ML, alongside explainability, privacy, accountability, and transparency. Neither the US nor the European Union (EU) provide any guidance specific to ML in clinical research, although the FDA provides guidelines for regulatory assessment of ML for clinical care \citep{fda}, and the EU provides guidance on the general development of AI \citep{Tiple2020RecommendationsOT}. With respect to clinical trials, international committees have recently published the SPIRIT-AI guidelines for trial protocols~\citep{Rivera2020GuidelinesFC} and the CONSORT-AI guidelines for trial reporting~\citep{Liu2020ReportingGF}. While these address the practical inclusion of AI/ML techniques in clinical research, they mention fairness only in passing. Accordingly, researchers have emphasized the need for explicit guidance and recommendations of minimum fairness standards from regulatory bodies~\citep{WawiraGichoya2021EquityIE}. We argue for greater consideration of issues of fairness in the development of ML for healthcare from ideation to implementation. 

\subsection{Fairness in Machine Learning}
The study of fairness in machine learning aims to ensure that ML models do not ``systematically and unfairly discriminate against certain individuals or groups of individuals in favor of others''~\citep{friedman1996bias}. Individuals or groups are typically characterized via certain (demographic) attributes associated with sub-populations who are vulnerable to harm due to structural biases~\citep{mehrabi}, such as age or race. These characteristics are legally termed \emph{sensitive} or \emph{protected} attributes~\citep{barocastext}. Here, we briefly review existing notions of fairness in situations where machine learning models inform or make decisions that affect the livelihood or well-being of humans and summarize challenges with these notions. We defer an introductory overview of existing fairness definitions and approaches to achieve them in practice to Appendix~\ref{sec:app:fairness}.

There are several popular perspectives seeking to understand how discrimination of data-driven decision-making can be formalized. Just like ethicists, political philosophers, legal scholars, and social scientists still grapple with what is fair, just, or equitable, there is an ongoing debate about the most reasonable formal definitions of fairness in ML and which ethical principles to consider~\citep{barocas,pmlr-v81-binns18a,holstein2019improving,friedler2021possibility,greene2019better}. 
In practice, often only statistical \textit{group fairness} notions can be operationalized well enough in a task-agnostic manner to suit real-world application. However, beyond incompatibility results between these definitions~\citep{chouldechova2017,kleinberg}--and consequently the difficulty of choosing one out of many group fairness definitions--a comprehensive empirical examination of the impact of penalizing group fairness violations on a variety of measures of model performance and group fairness metrics on various clinical datasets and prediction tasks found that such procedures lead to near-universal performance degradation along multiple dimensions. This indicates that group fairness measures are often not an effective solution for handling unfairness in healthcare settings~\citep{pfohl2021}.
On the other hand, \textit{individual fairness} suffers from the required normative choice of a ``similarity metric'' on individuals, which is difficult to formalize even in well-studied contexts.
While causal definitions are conceptually closest to legal notions of fairness~\citep{loftus}, there still remains fundamental disagreement about the most applicable concepts of fairness in particular settings~\citep{paulus,hu2020s}. Causal definitions require strong assumptions on the underlying data generating mechanism~\citep{loftus,kusner2017,kilbertus2017avoiding} and are thus difficult to operationalize. Since there is no universal technical solution to fairness in ML, the path forward is to involve domain experts, affected stakeholders, and to recognize social context, including the potential for institutionalized discrimination, and examine each situation holistically~\citep{Selbst2019FairnessAA,benjamin,pfohl2021}.

\section{Possible Sources of Unfairness in Clinical Trials}
\label{unfairness}
While ethical guidelines attempt to advise in a principled and humane approach towards clinical research, there still continues considerable debate regarding the construction of clinical trials. With respect to the seven principles of ethical clinical studies (Section~\ref{ethicalguidelines}), we delineate possible sources of unfairness to expose where further advancement of trial designs and implementation of fair ML can contribute towards more ethical clinical research. 

\subsection{Subject Participation}
\label{subjectparticipation}
Women, racial/ethnic minorities, and other under-served groups have been historically under-represented in CTs~\citep{Mccarthy1994HistoricalBO}. Disparities have been identified in the study of cardiovascular disease~\citep{cho2021increasing,Ranganathan2006ExclusionAI,Heiat2002RepresentationOT}, oncology~\citep{Duma2018RepresentationOM,Chen2014TwentyYP}, and acquired immunodeficiency syndrome (AIDS)~\citep{Shepherd1994WomenAH,Gwadz2009AnEB}; these particularly affect women of racial/ethnic minorities~\citep{Lippman2006TheIO,Gwadz2009AnEB}. While most CTs have been implemented in relatively wealthy regions such as North America and Western Europe~\citep{Thiers2008TrendsIT}, the pharmaceutical industry has driven globalization of clinical research, particularly to developing countries~\citep{Shah2003GlobalizationOC,Silva2016GlobalizationOC}. Although this may improve diversity, it also requires renewed consideration of ethics and fairness~\citep{Shah2003GlobalizationOC,Silva2016GlobalizationOC, Thiers2008TrendsIT}.

The value of inclusion has been widely demonstrated, with significant survival benefits~\citep{yang2019sex}. Disparities in medical knowledge have long been criticized, with many questioning how the ``white male [came] to be the prototype of the human research subject''~\citep{cotton1990examples,dresser1992wanted,camidge2021race}. This may result in under-studied groups receiving ineffective or harmful treatments (or none at all) limited access to hospital care~\cite{sjoding2020racial}. For example, women experience a notably higher incidence of adverse drug reaction and toxicity than men, but few trials explore sex differences~\citep{Zopf2008WomenEA,Zucker2020SexDI,Unger2022SexDI}. Under-representation can also affect statistical trial outcomes, diminishing knowledge generalizability~\citep{Rothwell1995CanOR,Kent2007LimitationsOA,Ioannidis1997TheIO,Weiss2008GeneralizabilityOT}. Unfair under-representation may occur during the process of subject selection, recruitment, and enrollment in a clinical trial. It may arise from explicit exclusion of certain subgroups (Section \ref{categoricalsubjectexclusion}), or under-representation caused by socio-economic disparities, discriminatory practices, or communication barriers (Section \ref{underrep}).

\subsubsection{Categorical Subject Exclusion} 
\label{categoricalsubjectexclusion}

International ethical guidelines for conducting human medical research have asserted that subjects of research should be selected such that burdens and benefits are equitably distributed and exclusion of groups should be justified~\citep{cioms, world1991declaration, Levine2002InternationalEG,Emanuel2000WhatMC}. However, studies have found that inclusion/exclusion criteria are often restrictive and unjustified in their exclusion of under-served groups~\citep{VanSpall2007EligibilityCO,Ranganathan2006ExclusionAI,Bodicoat2021PromotingII,KennedyMartin2015ALR}. Although some exclusion may be sensible (e.g., people without prostates from prostate cancer studies), researchers have identified explicitly codified exclusion of certain racial/ethnic groups~\citep{Ranganathan2006ExclusionAI}, women, the elderly, and those with co-morbidities without good reason~\citep{VanSpall2007EligibilityCO, Heiat2002RepresentationOT,KennedyMartin2015ALR}. Eligibility criteria may also indirectly exclude sub-populations; some studies excluded participants on the basis of language and birthplace~\citep{Ranganathan2006ExclusionAI} and others on the basis of co-morbidities that are more prevalent amongst certain racial/ethnic groups~\citep{WebbHooper2019ReasonsFE}. The inclusion of vulnerable individuals (e.g., pregnant women~\citep{Baylis2012ResearchIP}, children, or people from low-income countries) has required particular justification~\citep{Brady2014TheBR}, often leading to their exclusion. 

\begin{tcolorbox}
\textbf{Case 1: Participation of pregnant women in clinical research.}
      Pregnant and lactating women remain under-represented in clinical research, despite advocacy from international organizations~\citep{cioms}. Clinical testing of experimental vaccines has historically excluded pregnant women, due to potential harm to the fetus, physiological complexity of pregnancy, and legal or regulatory complications~\citep{Schwartz2018ClinicalTA}. Consequently, they and their infants are often unfairly deprived of protections against potentially catastrophic infections. For example, although generally asymptomatic, the Zika virus can produce congenital malformation syndrome and pediatric neuro-developmental abnormalities~\citep{Schwartz2017TheOA}; an effective Zika vaccine would be most beneficial to unborn infants. Vaccine-preventable infections are also a leading cause of morbidity in pregnant women~\citep{Faucette2015ImmunizationOP,Faucette2015MaternalVM}. Pregnant women have also been excluded from clinical trials during a recent Ebola outbreak in 2018~\citep{Breman2016DiscoveryAD,Schwartz2018ClinicalTA} and the current COVID-19 pandemic~\citep{Rubin2021PregnantPP, Taylor2020InclusionOP, McKay2020ProtectionBE}, despite mounting evidence that pregnant woman may have a higher risk of COVID-19 complications~\citep{Villar2021MaternalAN}.
\end{tcolorbox}

\subsubsection{Under-representation} 
\label{underrep}
Even in the absence of explicit categorical exclusion, under-representation may result from i) unequal access to care due to socio-economic and geographical differences, ii) implicit physician biases, and iii) communication issues between patients and investigators, among other causes~\citep{Nipp2019OvercomingBT,Nelson2002UnequalTC,Fisher2011ChallengingAA,Clark2019IncreasingDI}. Subgroups with historically lower financial resources are consistently under-represented in cancer clinical trials, likely owing to high costs of frequent travel to trial sites, inability to take time away from work, or caution against the financial impacts of adverse effects, compounding with already high routine care costs \citep{Nipp2019OvercomingBT,Clark2019IncreasingDI}. People of color have also historically had inadequate access to medical care; minorities are more likely to receive care at under-resourced hospitals \citep{HasnainWynia2007DisparitiesIH} and are less likely to have health insurance, which is often a pre-requisite for medical facilities and clinical trials \citep{Fisher2011ChallengingAA}.

Physicians are often responsible for determining who to approach for enrollment. Those who harbor biases may believe that minorities are less likely to remain adherent to treatment recommendations \citep{vanRyn2000TheEO,Sabin2008PhysicianIA}, or likely to reject trial offers \citep{Hamel2016BarriersTC}. However, research has shown that minorities are comparably willing to participate in clinical trials, but are approached far less \citep{Wendler2006AreRA}. Many clinical trials recruit on a first-come, first served basis; those with wealth, power, and connections are thus more likely to receive scarce resources, a particular challenge in unforeseen catastrophes such as the COVID-19 pandemic \citep{jansen2021fair}. Communication issues affecting economically and educationally disadvantaged individuals, such as language-related factors, lack of access to technology, or low levels of health literacy~\citep{stone2003invisible,denny2007clinical}, may also undermine fair subject selection, particularly in the absence of facilitating instruments (e.g., interpreters).

\begin{tcolorbox}
\textbf{Case 2: Women in AIDS studies.} Perhaps due to the historical stereotype that AIDS primarily afflicts gay men, its impact on women has been under-studied, with women excluded from studies or included in numbers too small to yield meaningful outcomes. Although infections are increasing most rapidly amongst women, the number participating in clinical trials of AIDS drugs, which provide an important source of first-rate medical care and access to experimental treatments, lags behind expectations~\citep{Shepherd1994WomenAH}. As a result, lower case survival rates have been observed in women diagnosed with AIDS. Studies involving women have focused on the reduction or prevention of transmission of human immunodeficiency virus (HIV) from a pregnant woman to a fetus or newborn, not on female-specific manifestations of the disease. Recent efforts have begun to address these concerns, arguing that what is known about the natural history of the diseases and their treatment in men may be inapplicable to women~\citep{Shepherd1994WomenAH}.
\end{tcolorbox}

\subsection{Treatment Allocation}
A critical component of a trial design is the mechanism by which participants are allocated to treatment arms. In the absence of scientific and ethical justification, randomized schemes are preferable in most circumstances, as they protect against allocation bias and facilitate equal baseline distribution of prognostic factors between the compared groups~\citep{paludan2016mechanisms}. As reported in the CIOMS International Ethical Guidelines, RCTs in particular equalize the foreseeable benefits and risks of participation~\citep{cioms} and also provide recognized scientific and statistical superiority~\citep{stolberg2004randomized,rosenberger2019randomization}. However, even in RCTs, issues of unfairness may still arise, either due to their inherent characteristics or to violations of ethical principles. 

\subsubsection{Randomized Controlled Trials}
The moral justification of clinical trials relies on the principle of \textit{clinical equipoise}, wherein researchers must possess ``genuine uncertainty'' regarding the benefits of new therapies over existing standard practices~\citep{Freedman1987EquipoiseAT,lilford1995equipoise}. However, RCTs typically investigate treatments that show promise in pilot trials, but are usually conducted with a fixed equal allocation ratio, uniformly distributing benefits and burden. While successful small-scale trials are not considered sufficient evidence of treatment effectiveness by regulatory agencies, as they often lack statistical power, they may influence an investigator's uncertainty. Preliminary evidence from pilot trials (and from uncontrolled trials~\citep{diaz2005pasteur}) may provide investigators with reason to believe in the success of an intervention, especially if the treatment effect is large, violating clinical equipoise. Adaptive designs with unequal~\citep{Meurer2012AdaptiveCT} or adaptive allocation ratios~\citep{avins1998can,peckham2015use} (based on treatment success or patient characteristics) may be more ethical alternatives, given the responsibility of practitioners to minimize the number of people given unsafe or inferior treatments (or placebos)~\citep{bhatt2016adaptive,pallmann2018adaptive}.

\begin{tcolorbox}
\textbf{Case 3: Ethics of RCTs.} The ACTG076 trial was conducted to examine whether zidovudine (AZT) could reduce vertical maternal-to-infant HIV transmission, compared against a placebo~\citep{connor1994reduction}. A conventional randomization design was used to ensure equal allocation of pregnant women, with 239 receiving AZT and 238 receiving placebo. However, 16 newborns in the AZT group and 52 in the placebo group had at least one HIV-positive culture, with a higher (>3x) HIV transmission in infants on placebo treatment. Subsequent researchers found that a suitable AD could have saved 11 newborns without significant loss of efficiency~\citep{yao1996play,biswas2001robust}.
\end{tcolorbox}

Clinical equipoise is used to justify the contradiction between physicians' therapeutic obligation to treat patients with the most beneficial possible methods and the true goal of clinical research: collecting evidence~\citep{Miller2003ACO}. This mindset has been criticized as contributing to \textit{therapeutic misconception}, where individuals believe that the primary goal of a clinical trial is to improve their outcomes, while the true goal of a trial is to produce experimentally-validated knowledge, where scientific validity is prioritized over patient outcomes~\citep{Henderson2007ClinicalTA,Meurer2012AdaptiveCT}. Adaptive designs may provide a way to improve patient outcomes within clinical trials, while maintaining scientific validity, thereby reducing the impacts of therapeutic misconception~\citep{Meurer2012AdaptiveCT}.

\subsubsection{Adaptive designs: benefits and ethical controversies}
\label{ads_benefits}
In some cases, the flexible and efficient nature of adaptive designs may offer a corrective to certain ethical difficulties of traditional RCTs~\citep{burnett2020adding,villar2021temptation}, resulting in more patient benefits~\citep{lee2015commentary,bhatt2016adaptive, pallmann2018adaptive}. However, adaptive designs generate their own ethical concerns~\citep{london2018learning} and may propagate inequalities. For example, clinical equipoise may be violated when data from the first group of study participants is evaluated~\citep{saxman2015ethical,laage2017ethical}. In RAR designs where allocation probabilities are skewed in favor of more promising treatment(s), it may be difficult to fully blind investigators to treatment arms. As enrollment progresses, investigators may recognize that more patients are being assigned to a certain arm and infer the superiority of the associated treatment~\citep{saxman2015ethical,laage2017ethical}. 
Sequential adaptation may also lead to different treatment of patients depending on time of enrollment~\citep{Legocki2015ClinicalTP}. The later a patient joins the study in relation to the target sample size, the higher their chance of receiving a superior  treatment under the changing randomization scheme, leading to an unfair distribution of risk and benefit across participants.

\begin{tcolorbox}
\textbf{Case 4: Benefits of Adaptive Designs.} 
\citet{Lee2020ACA} argue that equipoise may be violated during COVID-19 RCTs that study treatments that are expected to work. Following an interim analysis in the adaptive ACTT-1 trial for evaluating the experimental remdesivir treatment for COVID-19~\citep{Beigel2020RemdesivirFT}, which found that patients treated with remdesivir had a 31\% faster time to recovery than those who received placebo, the National Institute of Allergy and Infectious Diseases (NIAID) chose to offer patients in the placebo group the opportunity to receive remdesivir. Although this decision was sharply criticized, it was ethically justifiable, as at the time the NIAID chose to stop the trial, it was clear the treatment was more effective than placebo ~\citep{mozersky2020national}. Adaptive designs can expose fewer patients to ineffective and/or toxic interventions, ensuring a fairer distribution of superior treatments~\citep{Park2018CriticalCI}. 
\end{tcolorbox}

Trial adaptations based on interim results are not guaranteed to be beneficial, as both partial and final results may be inaccurate and biased in both magnitude and direction~\citep{villar2015,robertson2021point}. The sequential dependence induced by adaptive designs can lead to problems in confirmatory trials with frequentist errors controls~\citep{villar2015,deliu2021efficient}, exposing participants to inferior treatments and invalidating study results. Additionally, while adaptations should be pre-defined in a trial protocol, \citet{Park2018CriticalCI} caution that investigators may make changes post-hoc, based on results, introducing investigator-driven bias, although such issues could be mitigated with an independent monitoring committee. \citet{Legocki2015ClinicalTP} discuss concerns of informed consent due to the difficulties of clearly explaining the structure of adaptive designs to potential participants~\citep{Legocki2015ClinicalTP}. Adoption of adaptive designs is also hindered by perceived technical and logistical complexity~\citep{chow}; current works have sought to clarify methodology to improve accessibility to a wider audience~\citep{pallmann2018adaptive, robertson,chow}.

\subsection{Throughout a Trial}
Sources of unfairness may arise at trial planning or subject selection, but extend throughout trial implementation and generate post-trial consequences. 

\subsubsection{Differing Patient Risk Preferences}
\label{riskpref}
CT participants may operate with distinct risk preferences, defined in the economics and psychology literature as the ``extent to which people are willing to take on risk''~\citep{Charness2013ExperimentalME}. In clinical research, these are described as \textit{benefit-risk preferences}, where patients may find varying severity of side effects or consequences to be acceptable given potential benefits~\citep{BrettHauber2013QuantifyingBP,ReedJohnson2009MultipleSP,LlewellynThomas1991PatientsWT,Reed2021QuantifyingBP}. For example, a survey of older Americans found that they were willing to accept considerable risk--a 1-year risk of over $30\%$ of death or permanent severe disability from stroke--in exchange for an Alzheimer's disease treatment that would prevent disease progression beyond a mild stage~\citep{Hauber2009OlderAR}.

\begin{tcolorbox}
\textbf{Case 5: A study examining counseling for post-mastectomy breast cancer patients.}
An RCT sought to evaluate the effectiveness of counselling versus no counselling for participants undergoing mastectomy for breast cancer \citep{Brewin1989PatientPA}. However, willingness to engage in counselling would have had a significant impact on the overall success of counselling, confounding treatment effects. This would lead to invalid results; those who preferred to enter counselling may have felt improvements as a result, but those effects would be obfuscated by those who did want to engage in counselling and inevitably did not experience improvements. In response, \citet{Brewin1989PatientPA} propose a preference-based trial design, where patients are allocated to their treatment preferences (unless they have none, in which case they are randomized) to mitigate these confounders. 
\end{tcolorbox}

Researchers, regulators, and private companies increasingly agree upon the importance of incorporating patient preferences into clinical research~\citep{vanOverbeeke2019DesignCA,Soekhai2019MethodsFE,Stafinski2015IncorporatingPP}. Patient risk preferences provide insight into the types of treatments patients prefer, the consequences they may tolerate, and what they consider to be meaningful clinical benefit~\citep{Stafinski2015IncorporatingPP,Minion2016EndpointsIC}. Elicitation of risk preferences may aid in patient recruitment, as uncertainties regarding risk prevent some from participating in CTs~\citep{Nipp2019OvercomingBT}. For example, a study of perceptions of cardiovascular CTs found that women perceived greater harm from trial participation than men, contributing to their under-representation in cardiovascular CTs~\citep{Ding2007SexDI}.

Many patients also express a dislike of randomization, distrust of researchers, or strong treatment preferences~\citep{Unger2016TheRO,Torgerson1998UnderstandingCT}, causing them to refuse participation, negatively impacting the generalizability of a study to the outside population. Patients who agree to randomization but receive a non-preferred treatment may suffer resentful demoralization~\citep{Brewin1989PatientPA}, resulting in poor treatment compliance or dropout, or a negative placebo affect that compromises study outcomes~\citep{King2005ImpactOP,Torgerson1998UnderstandingCT,Brewin1989PatientPA}. In such a case, a preference-based study design may lead to better study outcomes due to improved adherence, and could be considered more individually ethical and equitable. However, abiding by patient preferences can also act as a confounder between the received treatment and the outcome, complicating statistical analysis.

\subsubsection{Patient Population Heterogeneity and Trial Generalization}
\label{heterogeneity}

Participants of a clinical trial are often heterogeneous, differing from each other as well as from the outside population. Patients suffering from the same general conditions may have diverse health characteristics and levels of health risk~\citep{Ioannidis1997TheIO, Kent2007LimitationsOA}. The prognoses of many diseases are highly variable. For example, COVID-19 ranges from non-symptomatic response to severe clinical courses requiring intensive care, sometimes leading to death~\citep{zhou2020clinical}. 
Heterogeneity may also be the result of several other factors, including biased subject selection practices (Section~\ref{subjectparticipation}) or differing patient risk preferences (Section~\ref{riskpref}). Patients who agree to enter clinical trials may be systematically distinct from those who refuse~\citep{LlewellynThomas1991PatientsWT,Nipp2019OvercomingBT, Torgerson1998UnderstandingCT}. Additionally, patient population drift, where the characteristics of the trial participant population change over time, may arise with continuous patient recruitment~\citep{Senn2013SevenMO}, or trial expansion to different participant populations~\citep{Villar2015PatientDA}.

Heterogeneity may impact the internal validity of a trial; if the treatment effect varies across subgroups, its estimate is only meaningful for a well-defined population, and can be misleading for differing individuals~\citep{stallard2020efficient}. Consequently, a minority of high-risk patients may have the majority of impact on trial outcomes~\citep{Ioannidis1997TheIO, Ioannidis1998HeterogeneityOT}. Heterogeneity of health risk may cause some patients to be unknowingly and unfairly susceptible to greater adverse effects of treatment~\citep{Kent2007LimitationsOA,Kent2008CompetingRA}. In an adaptive design, patients who enroll at different times may experience changed trial characteristics; participants may also be subject to trial modifications based upon preceding patients with different health characteristics~\citep{Legocki2015ClinicalTP}. 

Heterogeneity may also lead to problems with the external validity of a trial, such that the trial results may not apply to the general population, due to the specific characteristics of the trial participants~\citep{Rothwell1995CanOR,Kent2007LimitationsOA,Ioannidis1997TheIO,Weiss2008GeneralizabilityOT}. This negatively impacts not only the scientific validity of a trial by providing misleading results, but can also propagate health disparities as the impact of treatments on certain subgroups remain underexplored~\citep{Chen2014TwentyYP,Dhruva2008VariationsBC,Lippman2006TheIO,cotton1990examples}. 

\begin{tcolorbox}
\textbf{Case 6: Cardiovascular disease trials and Medicare patients.}
In the US, Medicare is a federal program that provides health insurance coverage for the elderly.
Cardiovascular disease is the leading cause of death in Medicare beneficiaries. The federal agency responsible for Medicare purports to make coverage decisions (funding of procedures and treatments) based upon evidence of improvement in health outcomes~\citep{Dhruva2008VariationsBC}, typically derived from CTs. Researchers found that the participants of cardiovascular/heart failure CTs that inform Medicare coverage decisions differ significantly from the population of Medicare beneficiaries~\citep{Dhruva2008VariationsBC,Masoudi2003MostHO}. Women and the elderly were significantly under-represented, with CT participants likely to be younger and male (avg. age 60.1, $75.4\%$ male) than Medicare beneficiaries (avg. age 74.7, $48.1\%$ male)~\citep{Dhruva2008VariationsBC}. Thus, decisions regarding the provision of treatments to patients (particularly elderly women) have been unfairly made on the basis of evidence that does not reflect the patient population~\citep{Dhruva2008VariationsBC}. 
\end{tcolorbox}

Trial design and analysis methods may mitigate the impact of patient heterogeneity, although further research and adoption is crucial~\citep{Kent2008CompetingRA}. Subgroup analyses explore whether treatment effects differ based upon certain patient characteristics~\citep{Pocock2002SubgroupAC, Gabler2016NoII}. In conventional subgroup analyses, participants are stratified using single patient characteristics, such as age; however, this type of analysis has been criticized as inadequately informative \cite{Rothwell1995CanOR,Kent2007LimitationsOA}. Risk-stratified subgroup analysis, where risk levels are determined by patient attributes, may be more representative~\citep{Kent2007LimitationsOA,Ioannidis1998HeterogeneityOT,Hayward2006MultivariableRP,Rothwell2005FromST}. Unfortunately, a recent study has found a significant decrease in the usage of appropriate methods for subgroup analyses in RCTs over time~\citep{Gabler2016NoII}. Prognostic or risk factors may also be used in alternative randomization schemes, such as stratified randomization, where patients are divided into subgroups and randomized to subgroup-specific treatment arms~\citep{Kernan1999StratifiedRF}, or novel adaptive enrichment designs, where recruitment is focuses on subgroups most likely to benefit~\citep{friede2020adaptive, pallmann2018adaptive,ondra2016methods}.

Temporal heterogeneity particularly impacts trials that span a long duration, such as studies of rare diseases, which have embraced the use of RAR~\citep{abrahamyan2016alternative}. In RAR designs,
unobserved time trends can inflate type-I error~\citep{simon2011using}, even if restricted randomization is used~\citep{rosenberger2015randomization}; temporal changes in trial data must therefore be carefully evaluated~\citep{villar2018response}. Even robust procedures~\citep{simon2011using} may be a computational burden and suffer form reductions in statistical power, indicating the need for computationally feasible testing procedures~\citep{villar2018response}, or adaptive strategies that may incorporate non-stationarity~\citep{robertson}.

\section{Opportunities for Machine Learning}
In this section, we explore opportunities for ML to mitigate sources of unfairness (Section~\ref{unfairness}), emphasizing the consideration of fairness and ethics during development of ML. For each topic, we describe the problem setting, the current literature, highlight gaps in knowledge, and provide recommendations for future directions. 

\subsection{Subject Selection}
\label{subjectselection}
Unfairness may arise throughout the process of subject selection, recruitment, and enrollment for clinical trials (Section~\ref{subjectparticipation}). A large portion of ML research for clinical trials has focused on two problem settings: i) evaluation of eligibility criteria, and ii) patient identification for clinical research or trial matching. We consider these alongside the issues that may arise if fairness and bias are not carefully considered during ML development.

\subsubsection{Evaluation of Eligibility Criteria}
Categorical subject exclusion, typically codified in trial eligibility criteria, often induces unfairness in subject selection. Recent work has sought to use ML for data-driven evaluation and development of trial eligibility criteria to improve trial outcomes~\citep{Kim2021TowardsCD,Liu2021EvaluatingEC}. \citet{Kim2021TowardsCD} use EHR data to evaluate frequently used COVID-19 trial eligibility criteria, identifying alternative thresholds for eligibility criteria that would increase the number of desired observed outcomes with fewer patients. \citet{Liu2021EvaluatingEC} use EHR data to evaluate the impact of eligibility criteria on cancer trial populations, finding they could broaden criteria without sacrificing trial efficacy. Such methods can broaden unnecessarily restrictive eligibility criteria and, in future work, potentially improve fairness by exposing whether certain sub-populations were made unjustifiably ineligible.

\subsubsection{Patient Identification and Phenotyping}

Data-driven ML methods may provide a path forward to mitigating under-representation in clinical trials through automation of subject selection. Researchers have used ML to identify patients eligible for clinical trials from EHR data, through matching of patient characteristics with trial criteria~\citep{Ni2015AutomatedCT,Zhang2017AutomatedCO,Hassanzadeh2020MatchingPT,Zhang2020DeepEnrollPM} or existing trial participants~\citep{Miotto2015CasebasedRU}. Others have sought to identify patients who are most likely to agree to participate in a trial, for more efficient use of recruitment resources~\citep{Ni2015WillTP,Vazquez2020UsingSM}. Recently, researchers have expressed enthusiasm at the possibility of ML patient-matching algorithms to improve the diversity of trial cohorts~\citep{Woo2019AnAB}.

Some research focuses on phenotyping algorithms, which aim to categorize patients based on health outcomes or disease states, as an intermediary to patient-trial matching. Practitioners have developed both supervised~\citep{Liao2015DevelopmentOP} and unsupervised~\citep{Glicksberg2018AutomatedDC} ML methods to learn disease phenotypes from EHR data, to identify representative cohorts that could benefit from proposed interventions. Phenotyping algorithms may also help in trials development by exposing subtypes of a disease to target for study. \citet{Li2015IdentificationOT} identify subtypes of Type 2 diabetes, a notably heterogeneous condition; this may inform treatment and research. Future work should focus upon developments to patient identification, phenotyping, and trial-matching algorithms to improve not only performance, but also fairness, explainability, and privacy.

\subsubsection{Fair Practices}
Exclusionary practices in clinical trials may bias resultant data that is used for downstream ML tasks; for example, researchers have sought to mitigate gender biases in word representations derived from trials data, which would be used for clinical prediction tasks~\citep{Agmon2022GendersensitiveWE}. EHR data also suffers from multiple sources of bias, including: i) \textit{sampling bias}, where certain subgroups are over- or under-represented in the data, ii) \textit{label bias}, where outcome variables have different meanings across subgroups, and iii) \textit{feature bias}, where predictor variables have different meanings across subgroups~\citep{paulus}. Beyond its structural complexity, EHR data is also often plagued by missing data, where the absence of certain variables can lead to biases~\citep{Vassy2018YieldAB,Weber2017BiasesIB}. Sampling bias may occur due to inequalities in access to care leading to less data collected from under-served populations~\citep{hing,chen2021}, or from research resulting from non-representative data cohorts~\citep{DAgostino2001ValidationOT}. Inequalities in the treatment of patients across subgroups may lead to label or feature bias, as outcomes and diagnostics may vary due to biases that impact quality of care~\citep{williams2015,paulus}. As racial/ethnic categories are social concepts that do not reliably inform genetic or clinical distinctions, they can be a source of feature bias; their use in clinical prediction algorithms may lead to negative impacts on marginalized populations~\citep{vyas}.

ML practitioners must be careful when manually selecting target outcomes for healthcare research. \citet{obermeyer} found that a widely used algorithm to predict clinical risk for referral to care management programs displayed significant bias against Black patients. This occurred because the developers of the model used healthcare costs as a proxy for health outcome. However, in part due to entrenched societal disparities, less money is spent on Black patients with comparable health to white patients. Here, the choice of inappropriate labels led to biased predictions. This is relevant to the previously discussed works that use patient agreement of participation in a trial as an outcome label~\citep{Ni2015WillTP,Vazquez2020UsingSM}; predictions based upon this label may have unintended biases as, for example, a patient can only agree to participate if they are invited. Unfortunately, in comparison to White patients, a disproportionately small proportion of racial/ethnic minorities are invited to participate in clinical trials, despite expressing a similar willingness to participate~\citep{Wendler2006AreRA}. These concerns also apply to phenotyping algorithms that rely on disease labels, given biases present during diagnosis. As interest and adoption of ML for efficient subject selection grows, researchers must assess their methods for sources of potential bias on all fronts, including problem selection, data collection, and model development. 

\subsection{Eliciting and Incorporating Patient Risk Preferences}
Below, we provide a review of the study of preference elicitation and describe how patient risk-benefit preferences (Section~\ref{riskpref}) may be incorporated into the design and development of clinical trials.

\subsubsection{Risk Preference Elicitation}
The study of preference elicitation--methods of collecting, determining, and modelling user preferences--varies across research fields. In the economics and psychology literature, preference elicitation is used to measure and evaluate the attitude of individuals regarding risk~\citep{Charness2013ExperimentalME}. These methods can be categorized as \textit{revealed preference methods}, which determine preferences based on behavioral data and \textit{stated preference methods}, which rely on individuals to express their preferences using survey-based methods~\citep{Ali2012OrdinalPE}. Revealed preferences may be considered the true preferences of individuals~\citep{Lambooij2015ConsistencyBS,Ali2012OrdinalPE}, whereas stated preferences are commonly critiqued for their hypothetical nature. However, correspondence between the two can be high. In a study of vaccination preferences, \citet{Lambooij2015ConsistencyBS} found that the stated preferences of $80\%$ of individuals matched their revealed preferences. 

In healthcare scenarios, individuals are usually asked to assess their risk attitudes towards treatments that have not yet occurred and that they have no prior experience with, precluding reliable revealed preference methods. Studies on patient health typically use stated preference methods to quantify patient \textit{benefit-risk preferences} towards medical interventions~\citep{Ali2012OrdinalPE,BrettHauber2013QuantifyingBP,Reed2021QuantifyingBP,Soekhai2019MethodsFE}. These methods can generally be categorized into two types: \textit{direct-elicitation} methods, where participants are explicitly asked to identify acceptable levels of risk/benefit, and \textit{conjoint-analysis} methods, where risk/benefit is assessed based on survey responses~\citep{BrettHauber2013QuantifyingBP}. A direct-elicitation method is the \textit{standard gamble} technique, wherein participants choose between a certain health state and a gamble that could result in either a better or worse outcome. The probabilities of the gamble are modified systematically until the participants are indifferent, allowing for estimation of the utility of the certain health state~\citep{Soekhai2019MethodsFE, BrettHauber2013QuantifyingBP}. A conjoint-analysis method is a \textit{discrete-choice} experiment, wherein participants select their preferred scenario from a set of hypothetical choice situations~\citep{Soekhai2019MethodsFE, BrettHauber2013QuantifyingBP}. 

Researchers found that patients' reported preferences (versus true preferences) of treatment options can be affected by elicitation methodology~\citep{Bowling2005YouDD}, such as different framing of the same information~\citep{Edwards2001UnderstandingRA}. Patients may also disagree on the associations between numerical values and verbal descriptions of risk, with one study reporting that different patients associated the phrase ``frequent'' within a range of $30-90\%$~\citep{Woloshin1994PatientsIO,Edwards2001UnderstandingRA}. Reliable quantification of patient risk preference is challenging; more research must be conducted to address the incongruence between reported and true preferences, and to determine the underlying mechanisms behind individuals' responses to elicitation methods~\citep{Holzmeister2019TheRE}.

In the ML literature, preference elicitation focuses on models that represent user preferences~\citep{Chen2004SurveyOP,Blum2003PreferenceEA,Haddawy2003PreferenceEV,Dragone2018ConstructivePE}, often in the context of recommender systems, which learn preferences interactively. Attempts to automate the preference elicitation process across large populations include artificial neural networks with domain dependent priors~\citep{Haddawy2003PreferenceEV}, Bayesian preference elicitation by means of pairwise comparison queries~\citep{Guo2010RealtimeMB}, or supervised ML to improve direct elicitation results by recovering latent preferences~\citep{Clithero2019SupervisedML}. These methods seek to correct stated preferences to more closely mimic unobserved revealed preferences. ML may also assist in combining stated preferences with clinical data (which may expose revealed preferences) to achieve better estimates of true preferences~\citep{BrettHauber2013QuantifyingBP}; work has been pursued along these lines for non-health applications~\citep{Ali2012OrdinalPE}.
Ultimately, such software based solutions must also consider aspects of human-computer interaction. Stated preferences can be affected by interface design~\citep{Pommeranz2011DesigningIF}, suggesting that active user involvement in the interface design may yield more consistent and representative findings~\citep{cai2019human,vanLeersum2020WhatMT}. In summary, computationally assisted preference elicitation is a delicate task that requires a multi-disciplinary perspective.

\subsubsection{Incorporating Risk-Benefit Preferences}
Elicited risk-benefit preferences may be used in different ways to support ethical treatment. Patient risk preferences are valuable in pre-trial clinical research; they provide information regarding treatment outcomes and clinical benefits that patients find most important, what side effects patients may be willing to cope with, and the types of treatments patients most prefer~\citep{Stafinski2015IncorporatingPP,Minion2016EndpointsIC,McQuellon1995PatientPF}. For example, a study of ovarian cancer patients found that patients preferred cure and extension of life over lessened symptoms from chemotherapy, exhibiting that patients prefer decreased mortality over decreased morbidity~\citep{Minion2016EndpointsIC}. \citet{Stafinski2015IncorporatingPP} compare patient and clinician views on the relative importance of treatment outcomes for cardiovascular conditions, finding differing risk-benefit thresholds. \citet{Sparano2019ClinicianreportedSA} find that clinicians' reporting of symptomatic adverse events is incongruous with patient reported outcomes. Understanding patient risk profiles and how they relate to perceived clinical benefit is essential for more efficient and ethical choice of what treatments to pursue and compare within a clinical trial. 

Trial designs that incorporate patient risk preferences into the treatment allocation mechanism have been proposed~\citep{Brewin1989PatientPA,King2005ImpactOP,Wasmann2019PartiallyRP,Torgerson1998UnderstandingCT}. In one design, patients that have strong preferences receive their treatment of choice, while others are randomized as usual~\citep{Brewin1989PatientPA,King2005ImpactOP,Torgerson1998UnderstandingCT}. In another method, patient risk preferences are elicited prior to treatment, and all participants are still randomized. This allows researchers to incorporate preference information into post-trial analysis~\citep{Torgerson1998UnderstandingCT}, improving external validity without affecting internal validity~\citep{King2005ImpactOP,Wasmann2019PartiallyRP}. Novel preference-informed trials continue to be explored~\citep{Ali2021ANP} and future work should also examine the ethics and statistical validity of preference-informed designs (given confounders) and how preferences may be incorporated into ML-driven adaptive designs.

Recent work in fairness in ML has sought to incorporate notions of preference into definitions of algorithmic fairness, resulting in classifier outcomes that meet certain preference guarantees~\citep{Kim2020PreferenceInformedF,Hossain2020DesigningFF,Ustun2019FairnessWH,Zafar2017FromPT}. These works take inspiration from the economics notion of \textit{envy-freeness}, asking that any group/individual prefers their treatment over that of others. \citet{Zafar2017FromPT} and \citet{Ustun2019FairnessWH} present fairness as a notion of envy-freeness among groups in the setting of decoupled classifiers, whereas~\citet{Hossain2020DesigningFF} impose fairness constraints using model outcomes. These works primarily deal with supervised ML models, where preference is defined as classifier performance metrics, such as accuracy. \citet{Kim2020PreferenceInformedF} incorporate individual outcome preferences, introducing a notion of preference-informed individual fairness that allows for deviations from similarity-based individual fairness, given alignment with individuals' preferences. Future work may address how preference-based fairness occurs in multi-armed bandit models, RL models, or in online learning, and incorporate specified risk preferences in online treatment allocation. 

\subsection{Modelling Adaptive Designs}
\label{modelling}

The slow uptake of ADs, despite their benefits (Section~\ref{ads_benefits}), has been attributed to lack of familiarity, concerns of how funding bodies or regulators may view them, and lack of clarity regarding how ADs may be planned, implemented, interpreted, and reported in practice~\citep{burnett2020adding,robertson,chow,pallmann2018adaptive}. This hesitancy is accompanied by a wariness against inclusion of ML in these designs, despite ongoing theoretical and methodological advancements~\citep{robertson,villar2015,williams2021,williamson,aziz2021multi,Bastani2015OnlineDW,Lee2021SDFBayesCO}. 

Since the 1930s, ML models have been proposed for use in adaptive designs. Thompson sampling, also known as Bayesian response adaptive randomization, was first proposed as a solution to patient treatment allocation in clinical trials~\citep{Thompson1933ONTL}. It is a solution strategy to the multi-armed bandit problem (MABP), where an agent trades off exploration (the acquisition of new information) with exploitation (optimal decisions based on existing knowledge). In a MABP, a fixed set of resources must be sequentially allocated between competing choices; in the trial setting, we may view the patients as the fixed set of resources and the trial treatment arms as the competing choices~\citep{villar2015}. Although the MABP seems to encapsulate in theory the practical problems of trials, there has been continual hesitance to adopt them~\citep{villar2015, Armitage1985TheSF}. The perceived shortcomings of bandit methods include loss of statistical power, challenges of statistical analysis (such as hypothesis testing) on bandit outcomes, and practical barriers to implementation (lack of access to statisticians or ML experts)~\citep{villar2015,robertson,williams2021}. 

Recent work has sought to innovate upon the field of multi-armed bandits (MABs) to support application to ADs, particularly RAR. \citet{villar2015} evaluate the performance of MABP approaches to RAR as compared to other allocation rules, including fixed randomization, finding that MABP methods improve patient welfare but reduce statistical power. The authors subsequently propose a solution building upon the Gittins index to improve statistical power~\citep{Villar2015ResponseadaptiveRF}. \citet{williamson} also propose a bandit-based design that aims to improve statistical power and minimizes treatment effect biases. \citet{williams2021} expose the difficulties of hypothesis testing on data derived from bandit models, and propose adjustments to existing statistical tests using knowledge of the bandit algorithm. 

Theoretical ML research has focused mainly on applying MABs to dose-finding~\citep{aziz2021multi, baek,Lee2021SDFBayesCO,Lee2020ContextualCL,shen2020learning,Bastani2015OnlineDW} or treatment allocation~\citep{Atan2019SequentialPR,Press2009BanditSP,Varatharajah2018ACA}. From the viewpoint of multiple (sequential) testing, \citet{yang2017framework,xu2021unified} propose a unified MABP based framework that allows for online false discovery rate (FDR) control, i.e., a pre-specified FDR is guaranteed at any time of the sequential tests (even when stopping adaptively), yielding good sample complexity, high statistical power, robustness to various types of distribution shift, and low FDR at any point during the sequential testing. More broadly, recently renewed interest in anytime-valid sequential hypothesis testing (e.g.,~\citep{lai1985asymptotically}) has surfaced practical methods for anytime-valid confidence intervals with FDR control even when adaptively combining or stopping sequences of statistical tests~\citep{howard2021time,grunwald2020safe,katsevich2020simultaneous,turner2021two}. Specifically for treatment effect estimation, \citet{Hadad2021ConfidenceIF} present a method based on adaptive reweighting under adaptively collected data that aims at high statistical power and asymptotically correct coverage. While such methods can maintain clear statistical guarantees for arbitrary adaptive designs, we expect that bridging the gap between ML theory and clinical trials practice remains challenging due to a lack of relevant expertise from trial investigators, requirement of multi-disciplinary teams, and reluctance from regulators to consider the relatively new and theoretically involved underlying statistical concepts. Opportunities for ML in the development of ADs include improved sequential MAB or other treatment allocation methods that i) maintain required statistical guarantees under adaptive changes to the trial, ii) incorporate fairness considerations (addressing patient heterogeneity), and iii) handle or mitigate the impact of patient time trends (addressing patient drift).

\subsection{Fair Exploration}
In Section~\ref{modelling}, we discussed the potential for explore-exploit methods (such as MABs or RL) to learn optimal policies by maximizing some notion of overall reward. Recent work has sought to characterize fairness in MABs~\citep{joseph,liu2017,Gillen2018OnlineLW, patil,chenbandit} and RL~\citep{jabbari}. These works evaluate fairness with respect to treatment arms/potential actions (rather than agents or users) from three general perspectives: i) \textit{meritocratic fairness}, where a worse arm should never have a higher chance of being selected over a better one~\citep{joseph}, ii) \textit{individual fairness}, where similar individuals (in this case, arms) should be treated similarly~\citep{liu2017,Gillen2018OnlineLW}, and iii) constraints on the proportions in which arms should be pulled~\citep{patil,chenbandit}. However, the nature of the explore-exploit trade-off leads to unaddressed fairness implications for CTs, namely that certain subgroups may be subject to a greater \textit{burden of exploration}, where they bear the negative consequences of exploration disproportionately~\citep{chouldechova}. The question of ``how to explore and randomize ethically'' is understudied from an ML perspective~\citep{chouldechova,kilbertus2020fair}. Exploration often requires algorithms to take actions that may ultimately be sub-optimal in order to gather information. In CTs this may correspond to the sacrifice of one individual's well-being for the collective benefit of others. If we view it as a moral obligation to take only optimal actions for studied individuals, at the expense of exploration, this may slow learning and lead to other types of unfairness due to a lack of information. Particularly in the clinical setting, exploration is at odds with the therapeutic obligation to treat patients as optimally as possible.

There has been comparatively sparse research on this conception of unfairness, which seeks to examine the burdens placed on certain subgroups, evaluating fairness from the perspective of the agents or users that are affected by the model outcomes, rather than the arms/actions. \citet{Jung2020QuantifyingTB} demonstrate the problem of ``free riding'' in a multi-agent setting, where ``free-riders'' can incur minimal regret by  accessing the information garnered by other agents. \citet{raghavan} introduce the concept of \textit{group externalities} to quantify the negative impacts the presence of one group may impose on another, under the linear contextual bandits model. \citet{baek} approach the problem in a similar fashion, introducing the concept of \textit{grouped bandits}, where groupings are defined by user characteristics and access to different subsets of arms. The authors develop a solution to learn fair policies through use of the \textit{Nash Social Welfare} (NSW) function, a notion of fairness in economics. In a slightly different approach, \citet{Hossain2020DesigningFF} use the NSW to achieve fairness in the multi-agent MAB setting, where individuals/groups are considered separate agents with separate reward distributions. Future work should investigate the compatibility of these methods with real-world adaptive trial settings and the performance of these methods with respect to optimal treatment and statistical power.

\subsection{Patient Heterogeneity and Study Generalization}
Patient heterogeneity in CTs may lead to reduced internal and external validity, as well as fairness (Section~\ref{heterogeneity}). ML can provide methods of identifying heterogeneity pre- and post-trial, assessing generalizability, and aiding with generalization once heterogeneity is exposed. Patient heterogeneity, owing to a multitude of characteristics, results in \textit{heterogeneity of treatment effects} (HTE), where treatment effects vary across individuals. In a clinical setting, the ultimate goal of HTE analysis is to estimate the causal effect for certain treatments on an individual level~\citep{Kent2018PersonalizedEB}.

\citet{Kent2018PersonalizedEB} identify two general approaches to predictive HTE analysis: i) \textit{risk modelling}, where a multivariate model that predicts outcome risk is used to stratify the patient population~\citep{Kent2007LimitationsOA,Hayward2006MultivariableRP,Imperial2021PrecisionEnhancingRS}, and ii) \textit{effect modelling}, where treatment effects are predicted directly, sometimes incidentally identifying subgroups and predictive covariates~\citep{Su2009SubgroupAV,Athey2016RecursivePF,Wager2018EstimationAI}. While multivariate risk-stratified analysis is effective for identifying HTE~\citep{Kent2007LimitationsOA,Hayward2006MultivariableRP}, it may suffer in the presence of other dimensions of risk, such as treatment-related harms, particularly if they are correlated with outcome risk~\citep{Kent2018PersonalizedEB}. Effect modelling can be better suited to account for heterogeneous risks or more individualized impacts on treatment effects~\citep{Kent2018PersonalizedEB}. Future work may focus on how ML can be practically incorporated into trial analyses; for example, \citet{Watson2020MachineLA} provide guidance on how ML can be used to determine HTE with strict type-I error control.

Other methods evaluate the generalizability of a clinical trial by assessing the representativeness of the trial population, comparing eligibility criteria with patients identified from large EHR datasets~\citep{Li2019AssessingTV,Sen2017CorrelatingEC}. \citet{Qi2021QuantifyingRI} assess representativeness using ML fairness metrics to identify under-represented subgroups. Such methods would benefit from further research on the interplay between representativeness, fairness, and generalizability. 

Generalization of ML models to new populations/settings and dataset drift over time have been highlighted as broad challenges facing practical adoption of ML for healthcare~\citep{Kelly2019KeyCF}. Acquiring useful data for ML is complicated by noisy data collection processes and differences in formats of EHR and other clinical data~\citep{Dexter2020GeneralizationOM}, which could be standardized. Generalization may be improved through existing methods, including use of  external validation datasets~\citep{Debray2015ANF} or multiple data sites~\citep{Nakano2020EnhancingMG}, and through identification and manipulation of covariates that impact generalizability~\citep{Futoma2021GeneralizationIC}. ML may also assist in identification of dataset drift, and provide recommendations of appropriate model-updating procedures~\citep{Davis2019ANU}, which can then be resolved through other methods, such as periodic model validation and manual model re-training. Future research may focus on further automation of dataset drift correction or advancements in continual learning, i.e., study of ML systems that continually learn and evolve based on new data~\citep{Futoma2020TheMO}. However, \citet{Futoma2020TheMO} argue that overly generalized ML models may not be clinically useful. They assert that practitioners should focus on ensuring that the methodological process of model development is generalizable and that a ML model is well-understood---this would allow for practitioners to modify a model accordingly given a new setting or population. This holistic viewpoint may provide better clinical utility, helping practitioners use ML technology both correctly and carefully.

\section{Conclusion}
ML may improve the efficacy and fairness of CTs, given careful implementation and focused advancements in the field. Prior to more widespread adoption of ML for CTs, it is vital that researchers and practitioners consider both the possible benefits and possible harms the introduction of new technology can bring. The wider scientific community relies upon critical information derived from CTs to serve as ground truths for medical research. While we recognize the benefits of rapid discovery, CTs should be as representative and valid as possible, such that these ground truths are either applicable to all, not just a privileged minority, or at least come with explicit caveats. In the development of ML for clinical trials, ethics and fairness are critical considerations alongside any others such as performance, efficiency, or statistical validity, and should be recognized as key from problem selection to implementation in practice. For these reasons, we have provided a comprehensive overview of the sources of unfairness, as well as the accompanying opportunities for ML research that can be pursued, with fairness firmly in mind, to mitigate these issues. We caution that any such pursuits should be overseen by a multi-disciplinary team that can thoroughly evaluate the goals and consequences of such work.

\begin{acks}
Isabel Chien and  Richard E. Turner are supported by an EPSRC Prosperity Partnership EP/T005386/1 between Microsoft Research and the University of Cambridge. Nina Deliu was supported by the NIHR Cambridge Biomedical Research Centre  (BRC-1215-20014). Sofia S. Villar thanks the UK Medical Research Council (grant number: MC\_UU\_00002/15). Adrian Weller acknowledges support from a Turing AI Fellowship under EPSRC grant EP/V025279/1, The Alan Turing Institute, and the Leverhulme Trust via CFI.
\end{acks}

\bibliographystyle{ACM-Reference-Format}
\bibliography{references}

\newpage
\appendix
\section{Seven principles for ethical clinical trials}
\label{sec:app:ethicalprinciples}

In discussions of ethical guidelines for clinical trials, three documents have been cited as universally influential~\citep{nihethics,Emanuel2000WhatMC,Nardini2014TheEO}: the \textit{Nuremberg Code} (1947)~\citep{code1949nuremberg}, the \textit{Declaration of Helsinki} (1964)~\citep{world1991declaration}, and the \textit{Belmont Report} (1979)~\citep{Brady2014TheBR}, which were all developed following egregious instances of patient abuse. The Nuremberg Code is a response to the atrocities committed by Nazi doctors during World War II. It establishes the need for informed patient consent and a favorable risk-benefit ratio~\citep{code1949nuremberg}. The Declaration of Helsinki was developed to supplement the Nuremberg Code and focuses upon the conduct between physicians and patients during research, particularly with respect to favorable risk-benefit ratio and independent review~\citep{Emanuel2000WhatMC,Nardini2014TheEO}.

Other notable documents~\citep{nihethics} include the CIOMS (Council for International Organizations of Medical Sciences) \textit{International Ethical Guidelines for Biomedical Research Involving Human Subjects} (2002)~\citep{cioms} and the \textit{U.S.\ Common Rule} (1981)~\citep{korenman2006teaching}. As these guidelines were created in reaction to particular circumstances, they have been criticized as lacking generality and even being in conflict with one another~\citep{Emanuel2000WhatMC,Christakis1991ExistingIE}. \citet{Emanuel2000WhatMC} have sought to integrate the principles discussed by these guidelines into a unified framework and argue that these principles apply universally and are consistent with philosophies of ``how reasonable people would want to be treated''.
Below, we detail the seven requirements developed by \citet{Emanuel2000WhatMC}.

\begin{enumerate}[leftmargin=*,topsep=0pt,itemsep=0pt]
    \item \textbf{Value}: A research study must provide value by contributing useful knowledge to society and improving health and well-being. This requires that the research also be reliable, generalizable, and widely shared. 
    \item \textbf{Scientific validity}: Research must be conducted in a ``methodologically rigorous manner''~\citep{Emanuel2000WhatMC}, using widely accepted methods, principles, and practices, and be practically feasible. Trials that compare different treatments must abide by the principle of \textit{clinical equipoise}, wherein researchers must possess ``genuine uncertainty'' regarding the benefits of new therapies over existing standard practices~\citep{Freedman1987EquipoiseAT,lilford1995equipoise}.
    \item \textbf{Fair subject selection}: Participants should be selected for trials on the basis of the scientific goals of the study, rather than unrelated traits, such as vulnerability or privilege. Equally, certain groups or individuals may not be excluded without valid scientific reason or excessive risk. Those who may incur risk or burden of the research should be able to benefit, and those who may benefit should take on some of the risks or burdens.
    \item \textbf{Favorable risk-benefit ratio}: In clinical research, possible risks should be minimized, potential benefits should be maximized, and the potential benefits to study participants and to society should either be proportional to or outweigh the possible risks.
    \item \textbf{Independent review}: To minimize biases or competing interests, proposals for clinical research should be evaluated by independent bodies who are not affiliated with or influenced by those conducting the study. 
    \item \textbf{Informed consent}: Individuals should be able to make an independent and well-informed decision, consistent with their own values and preferences, regarding whether they want to participate in clinical research. To achieve this, individuals should i) be ``accurately informed of the purpose, methods, risks, benefits, and alternatives to the research'', ii) understand the provided information and its relevance to their own circumstances, and iii) be able to make a ``voluntary and uncoerced'' decision regarding whether or not to participate~\citep{Emanuel2000WhatMC}.
    \item \textbf{Respect for enrolled subjects}: Individuals should be treated with respect throughout the entirety of the clinical research process, beginning from recruitment and extending until after participation. This includes, but is not limited to the following: i) respecting their privacy, ii) respecting the right for them to change their minds and withdraw from participation, iii) informing participants of any new information regarding the interventions or their own clinical conditions, iv) monitoring their well-being and ensuring proper treatment throughout the trial process, and v) informing them of the results and impacts of the clinical research. 
\end{enumerate}
We note that the discussion of ethics surrounding clinical trials extends beyond the debate over how to conduct an ethical trial. Scholars continue to contemplate the intrinsic morality of the practice of clinical trials as a form of experimentation that is simultaneously viewed by participants as an opportunity for treatment. This is often without the full understanding of the research subjects, who may view clinical trials as their best or only opportunity for care \citep{Timmermans2009ClinicalTA,stevenepstein}.

\section{Background on Fairness in Machine Learning}
\label{sec:app:fairness}

Anti-discrimination laws in the United States have provided two well-known interpretations for unfairness: \textit{disparate treatment} and \textit{disparate impact}~\citep{barocas}. \textit{Disparate treatment} refers to the different treatment, with intent, of similar people (with respect to non-sensitive attributes, but different sensitive attributes), while \textit{disparate impact} occurs with policies or practices that appear neutral but have a disproportionately adverse impact on those with certain sensitive attributes~\citep{barocas}. These notions lead to fairly straightforward translations into statistical definitions of group fairness.

\subsection{Definitions of fairness}

\textbf{Group Fairness} mostly measures the parity of a statistical measure (usually depending on the model outcomes and true outcomes) across all subgroups with different protected attributes~\citep{chouldechova}. For example, in the case of binary decisions, one can ask for equal rates of positive outcomes unconditionally of the true outcome (\emph{demographic/statistical parity} or \emph{equal allocation}). Conditioning on the true outcome yields a variety of definitions for so-called \textit{classification parity}, for example equal rates of errors such as false positive and negative rates (\emph{equal opportunity}, \emph{equalized odds}, \emph{disparate mistreatment})~\citep{hardt,zafar2017}. Conditioning on the predicted score instead, one can analogously define \textit{calibration parity} by asking individuals from different groups with the same predicted score to have the same probability of actually achieving a positive outcome~\citep{corbett,kleinberg,zafar2017}. Generally, all combinations of false positive, negative, discovery, omission rates as well as positive or negative predictive values can be considered as meaningful fairness definitions~\citep{zafar2019fairness}. Such statistical group fairness notions are popular because they are straightforward to interpret and often to achieve during model training without making additional assumptions about the data generating process. However, a drawback is that they do not give meaningful guarantees to individuals, structured subgroups, or intersections of protected groups, but only to ``average'' members~\citep{chouldechova}. In addition, most subsets of this collection of definitions cannot be satisfied simultaneously~\citep{chouldechova2017,kleinberg}, leaving open the question of which criterion to choose. Instead, interventional notions of fairness take into consideration the causal structure, i.e., how the protected attribute may have influenced other features, and asks for equal decisions ``had the protected attribute been fixed externally keeping everything else equal''~\citep{kilbertus2017avoiding}. Due to the assumption that the causal model~\citep{pearl} is known and ontological difficulties with attributing causal powers to variables such as ``race'' or ``gender'', these definitions are near impossible to operationalize~\citep{hu2020s}.

\citet{rajkomar} propose three principles inspired by distributive justice~\citep{Rawls1971ATO} specifically considering health equity that resemble statistical group fairness notions: i) \textit{equal patient outcomes} (when all subgroups receive equal benefit from the model),  ii) \textit{equal performance} (when a model is equally accurate for all subgroups), and  iii) \textit{equal allocation} (when resources are proportionately allocated to patients in all subgroups). \emph{Equal patient outcomes} is difficult to analyze as downstream effects of decisions may be unpredictable. While \emph{equal performance} and \emph{equal allocation} can be formalized as statistical group fairness criteria, they do not necessarily translate to \emph{equal outcomes}. While \emph{equal allocation} (statistical parity) is considered a crude criterion that is entirely blind to true outcomes, it can be relevant in healthcare settings. For example, historically, African American women with chest pain were sent for cardiac catheterization treatment at a lower rate in comparison with white men~\citep{schulman}, such that \textit{equal accuracy} would still propagate inequality, as these patients would be under-identified.

\textbf{Individual fairness} attempts to capture injustice experienced by a single individual compared to other (similar) individuals with different protected attributes. \citet{dwork} accordingly suggest that an algorithm is fair if similar individuals, according to a task-specific metric on the inputs, receive similar (distributions over) outcomes, again according to a fixed metric on outcomes. The main drawback is that the choice of similarity metrics is non-trivial in that they ultimately have to capture in which regards people should be considered similar/equal~\citep{chouldechova}. Despite attempts to learn such a metric from various types of online feedback \citep{Gillen2018OnlineLW,ilvento2019metric,bechavod2020metric,mukherjee2020two}, individual fairness remains difficult to implement in practice. \textit{Counterfactual fairness} may be viewed as a form of individual fairness that defines similarity with respect to a causal model of all measured features and protected attributes~\citep{kusner2017}. The underlying idea is that an individual should receive the same outcome in a counterfactual world in which they had a different protected attribute all else being equal. Again, due to the strong assumption of knowing the causal model, counterfactual fairness suffers from the same drawbacks as interventional fairness.

\xhdr{Another common definition} that fits to neither of the two categories is \textit{fairness through unawareness} (or \textit{anti-classification}), wherein an algorithm is fair if it does not consider sensitive attributes during its decision making process~\citep{mehrabi,corbett}. However, this practice has shown to be ineffective and potentially harmful~\citep{barocastext,corbett}.

\subsection{Approaches to achieve fairness}
Techniques to achieve fairness in ML models are commonly categorized by when the intervention occurs in the model building pipeline: pre-processing, at training time, and post-processing~\citep{barocastext,mehrabi}.

\xhdr{Pre-processing}
In the pre-processing approach, practitioners seek to transform the feature space into a representation that is independent of the sensitive attribute. This approach is agnostic to the downstream tasks that accept the representation as input. Correcting data biases is a difficult task, as it requires an understanding of how the measurement process is biased or intuition about how the data would appear in an ``unbiased'' setting~\citep{chouldechova}. Recent work has proposed that model prediction error can be decomposed in terms of bias, variance, and noise, and that these values can be used to inform additional data collection~\citep{chen2018}. Practically, learning fair representations is predominantly based on two potentially competing objectives: maintain information in the inputs that is relevant for accurate decision-making while removing all information about the protected attribute of a given input~\citep{zemel,feldman,madras,zhang}.

\xhdr{At Training Time}
At training time, fairness can be included as a constraint on the loss minimization. Achieving independence from the sensitive attributes at training time is beneficial as the classifier can be optimized with the specific fairness criterion in mind; a drawback is that we require access to the model training pipeline, and the final approach is usually model- or problem-specific~\citep{barocastext}. Typically, this approach involves either changes of the objective function or imposing a constraint based on the previously discussed fairness definitions~\citep{mehrabi,barocastext,zafar}.

\xhdr{Post-Processing}
In post-processing, a practitioners adjusts (the outputs of) a trained model so that it achieves a desired fairness criterion~\citep{barocastext,mehrabi}. Post-processing is most useful when one does not have access to the training data or learning algorithm. \citet{hardt} introduce a method to achieve equalized odds or equality of opportunity by (stochastically) changing certain model outputs. Similarly, \citet{kim2019} propose a framework to audit and post-process trained models to ensure accurate predictions across specified population subgroups.

\end{document}